# Risk-Sensitive Reinforcement Learning Applied to Control under Constraints

**Peter Geibel**                                                          PGEIBEL@UOS.DE
*Institute of Cognitive Science, AI Group*
*University of Osnabrück, Germany*

**Fritz Wysotzki**                                                   WYSOTZKI@CS.TU-BERLIN.DE
*Faculty of Electrical Engineering and Computer Science, AI Group*
*TU Berlin, Germany*

## Abstract

In this paper, we consider Markov Decision Processes (MDPs) with error states. Error states are those states entering which is undesirable or dangerous. We define the risk with respect to a policy as the probability of entering such a state when the policy is pursued. We consider the problem of finding good policies whose risk is smaller than some user-specified threshold, and formalize it as a constrained MDP with two criteria. The first criterion corresponds to the value function originally given. We will show that the risk can be formulated as a second criterion function based on a cumulative return, whose definition is independent of the original value function. We present a model free, heuristic reinforcement learning algorithm that aims at finding good deterministic policies. It is based on weighting the original value function and the risk. The weight parameter is adapted in order to find a feasible solution for the constrained problem that has a good performance with respect to the value function. The algorithm was successfully applied to the control of a feed tank with stochastic inflows that lies upstream of a distillation column. This control task was originally formulated as an optimal control problem with chance constraints, and it was solved under certain assumptions on the model to obtain an optimal solution. The power of our learning algorithm is that it can be used even when some of these restrictive assumptions are relaxed.

## 1. Introduction

Reinforcement Learning, as a research area, provides a range of techniques that are applicable to difficult nonlinear or stochastic control problems (see e.g. Sutton & Barto, 1998; Bertsekas & Tsitsiklis, 1996). In reinforcement learning (RL) an agent is considered that learns to control a process. The agent is able to perceive the state of the process, and it acts in order to maximize the cumulative return that is based on a real valued reward signal. Often, experiences with the process are used to improve the agent's policy instead of a previously given analytical model.

The notion of *risk* in RL is related to the fact, that even an optimal policy may perform poorly in some cases due to the stochastic nature of the problem. Most risk-sensitive RL approaches are concerned with the variance of the return, or with its worst outcomes, (e.g. Coraluppi & Marcus, 1999; Heger, 1994; Neuneier & Mihatsch, 1999), see also the discussion in section 3. We take the alternative view of risk defined by Geibel (2001) that is not concerned with the variability of the return, but with the occurrence of errors or undesirable states in the underlying Markov Decision Process (MDP). This means that we address a different class of problems compared to approaches referring to the variability of the return.

In this paper, we consider constrained MDPs with two criteria – the usual value function and the risk as a second value function. The value is to be optimized while the risk must remain below some specified threshold. We describe a heuristic algorithm based on a weighted formulation that finds a feasible policy for the original constrained problem.

In order to offer some insight in the behavior of the algorithm, we investigate the application of the algorithm to a simple grid world problem with a discounted criterion function. We then apply the algorithm to a stochastic optimal control problem with continuous states, where the set of feasible solutions is restricted by a constraint that is required to hold with a certain probability, thus demonstrating the practical applicability of our approach. We consider the control of a feed tank that lies upstream of a distillation column with respect to two objectives: (1) the outflow of the tank is required to stay close to a specified value in order to ensure the optimal operation of the distillation column, and (2) the tank level and substance concentrations are required to remain within specified intervals, with a certain admissible chance of constraint violation.

Li, Wendt, Arellano-Garcia, and Wozny (2002) formulate the problem as a quadratic program with chance constraints[1] (e.g. Kall & Wallace, 1994), which is relaxed to a nonlinear program for the case of Gaussian distributions for the random input variables and systems whose dynamics is given by linear equations. The nonlinear program is solved through sequential quadratic programming.

Note that the approach of Li et al. involves the simulation based estimation of the gradients of the chance constraints (Li et al., 2002, p. 1201). Like Q-learning (Watkins, 1989; Watkins & Dayan, 1992; Sutton & Barto, 1998), our learning algorithm is based on simulating episodes and estimating value and risk of states, which for the tank control task correspond to a measure for the deviation from the optimal outflow and the probability of constraint violation, respectively.

In contrast to the approach of Li et al. (2002), our RL algorithm is applicable to systems with continuous state spaces, whose system dynamics is governed by nonlinear equations and involve randomization or noise with arbitrary distributions for the random variables, for it makes no prior assumptions of either aspect. This it not a special property of our learning algorithm, and also holds true for e.g. Q-learning and other RL algorithms. The convergence of Q-learning combined with function approximation techniques necessary for continuous state spaces cannot be guaranteed in general (e.g. Sutton & Barto, 1998). The same holds true for our algorithm. Nevertheless, RL algorithms were successfully applied to many difficult problems with continuous state spaces and nonlinear dynamics (see e.g. Sutton & Barto, 1998; Crites & Barto, 1998; Smart & Kaelbling, 2002; Stephan, Debes, Gross, Wintrich, & Wintrich, 2001).

---

1. A constraint can be seen as a relation on the domains of variables restricting their possible values. If the variables in a constraint $C = C(x_1, \ldots, x_n)$ are random, the constraint will hold with a certain probability. Chance constrained programming is a particular approach to stochastic programming that considers constrained optimization problems containing random variables for which so-called chance constraints of the form $\mathbf{P}(C) \geq p$ with $p \in [0, 1]$ are formulated.

This article is organized as follows. In section 2, the RL framework is described. Section 3 reviews related work on risk-sensitive approaches. Section 4 describes our approach to risk-sensitive RL. In section 5, we elucidate a heuristic learning algorithm for solving the constrained problem using a weighted formulation. In section 6, we describe its application to a grid world problem. The tank control task is described in section 7. In section 8, experiments with the feed tank control are described. Section 9 concludes with a short summary and an outlook.

## 2. The RL Framework

In RL one considers an agent that interacts with a process that is to be controlled. At each discrete time-step, the agent observes the state $x$ and takes an action $u$ that in general depends on $x$. The action of the agent causes the environment to change its state to $x'$ according to the probability $p_{x,u}(x')$. Until section 7, we will consider the set of states, $X$, to be a finite set.

The action set of the agent is assumed to be finite, and it is allowed to depend on the current state. In each state $x$, the agent uses an action from the set $U(x)$ of possible actions. After taking action $u \in U(x)$, the agent receives a real valued reinforcement signal $r_{x,u}(x')$ that depends on the action taken and the successor state $x'$. In the case of a random reward signal, $r_{x,u}(x')$ corresponds to its expected value. The *Markov property* of an MDP requires that the probability distribution on the successor states and the one on the rewards depend on the current state and action only. The distributions do not change when additional information on past states, actions and rewards is considered, i.e. they are independent of the path leading to the current state.

The aim of the agent is to find a policy $\pi$ for selecting actions that maximizes the cumulative reward, called the return. The return is defined as

$$R = \sum_{t=0}^{\infty} \gamma^t r_t \,, \tag{1}$$

where the random variable $r_t$ denotes the reward occurring in the $t$-th time step when the agent uses policy $\pi$. Let $x_0, x_1, x_2, \dots$ denote the corresponding probabilistic sequence of states, and $u_i$ the sequence of actions chosen according to policy $\pi$.

The constant $\gamma \in [0, 1]$ is a discount factor that allows to control the influence of future rewards. The expectation of the return,

$$V^\pi(x) = \mathbf{E}\left[R \,|\, x_0 = x\right], \tag{2}$$

is defined as the **value** of $x$ with respect to $\pi$. It is well-known that there exist stationary deterministic policies $\pi^*$ for which $V^{\pi^*}(x)$ is optimal (maximal) for *every* state $x$. A stationary deterministic policy is a function that maps states to actions and is particularly defined as being independent of time and *Markovian* (independent of history). In this work, we will use the term of *maximum-value* policies instead of *optimal* policies just to distinguish them from minimum-risk policies that are also optimal in some sense, see section 4.1.

As usual, we define the state/action value function

$$Q^\pi(x, u) = \mathbf{E}\left[r_0 + \gamma V^\pi(x_1) \,\Big|\, x_0 = x, u_0 = u\right]. \tag{3}$$

$Q^\pi(x, u)$ is the expected return when the agent first chooses action $u$, and acts according to $\pi$ in subsequent time steps. From the optimal Q-function $Q^*$, optimal policies $\pi^*$ and the unique optimal values $V^*$ are derived by $\pi^*(x) \in \mathrm{argmax}_u Q^*(x, u)$ and $V^*(x) = Q^*(x, \pi^*(x))$. $Q^*$ can be computed by using Watkin's Q-learning algorithm.

In RL one in general distinguishes episodic and continuing tasks that can be treated in the same framework (see e.g. Sutton & Barto, 1998). In episodic tasks, the agent may reach some terminal or absorbing state at same time $t'$. After reaching the absorbing state, the agent stays there and executes some dummy action. The reward is defined by $r_t = 0$ for $t \geq t'$. During learning the agent is "restarted" according to some distribution on the initial states after it has reached the absorbing state.

## 3. Related Work

The random variable $R = \sum_{t=0}^{\infty} \gamma^t r_t$ (return) used to define the value of a state possesses a certain **variance**. Most risk-averse approaches in dynamic programming (DP) and reinforcement learning are concerned with the variance of $R$, or with its worst outcomes. An example of such an approach is *worst case control* (e.g. Coraluppi & Marcus, 1999; Heger, 1994), where the worst possible outcome of $R$ is to be optimized. In risk-sensitive control based on the use of *exponential utility functions* (e.g. Liu, Goodwin, & Koenig, 2003a; Koenig & Simmons, 1994; Liu, Goodwin, & Koenig, 2003b; Borkar, 2002), the return $R$ is transformed so as to reflect a subjective measure of utility. Instead of maximizing the expected value of $R$, now the objective is to maximize e.g. $U = \beta^{-1} \log \mathbf{E}(e^{\beta R})$, where $\beta$ is a parameter and $R$ is the usual return. It can be shown that depending on the parameter $\beta$, policies with a high variance $\mathbf{V}(R)$ are penalized ($\beta < 0$) or enforced ($\beta > 0$). The $\alpha$-*value-criterion* introduced by Heger (1994) can be seen as an extension of worst case control where bad outcomes of a policy that occur with a probability less than $\alpha$ are neglected.

Neuneier and Mihatsch (1999) give a model- free RL algorithm which is based on a parameterized transformation of the temporal difference errors occurring (see also Mihatsch & Neuneier, 2002). The parameter of the transformation allows to "switch" between risk-averse and risk-seeking policies. The influence of the parameter on the value function cannot be expressed explicitly.

Our view of risk is not concerned with the variance of the return or its worst possible outcomes, but instead with the fact that processes generally possess dangerous or undesirable states. Think of a chemical plant where temperature or pressure exceeding some threshold may cause the plant to explode. When controlling such a plant, the return corresponds to the plant's yield. But it seems inappropriate to let the return also reflect the cost of the explosion, e.g. when human lives are affected.

In this work, we consider processes that have such undesirable terminal states. A seemingly straightforward way to handle these *error states* of the system is to provide high negative rewards when the systems enters an error state. An optimal policy will then avoid the error states in general. A drawback of the approach is the fact that it is unknown how large the risk (probability) of entering an error state is. Moreover, we may want to provide a threshold $\omega$ for the probability of entering an error state that must not be exceeded by the agent's policy. In general, it is impossible to completely avoid error states, but the risk should be controllable to some extend. More precisely, if the agent is placed in some state $x$, then it should follow a policy whose risk *is constrained by* $\omega$. The parameter $\omega \in [0, 1]$ reflects the agent's risk-averseness. That is, our goal is not the minimization of the risk, but the maximization of $V^\pi$ while the risk is kept below the threshold $\omega$.

Markowitz (1952) considers the combination of different criteria with equal discount factors in the context of portfolio selection. The risk of the selected portfolio is related to the variance of the combined (weighted) criteria. Markowitz introduces the notion of the $(E, V)$-space. Our notion of risk is not related to the variance of $V$, but depends on the occurrence of error states in the MDP. Therefore the risk is conceptually independent of $V$, see e.g. the tank control problem described in section 7.

The idea of weighting return and risk (Markowitz, 1959; Freund, 1956; Heger, 1994) leads to the *expected-value-minus-variance-criterion*, $\mathrm{E}(R) - k\mathbf{V}(R)$, where $k$ is a parameter. We use this idea for computing a feasible policy for the problem of finding a good policy that has a constrained risk (in regard to the probability of entering an error state): value and risk are weighted using a weight $\xi$ for the value and weight $-1$ for the risk. The value of $\xi$ is increased, giving the value more weight compared to the risk, until the risk of a state becomes larger than the user-specified threshold $\omega$.

In considering an ordering relation for tuples of values, our learning algorithm for a fixed value of $\xi$ is also related to the ARTDP approach by Gabor, Kalmar, and Szepesvari (1998). In their article, Gabor et al. additionally propose a recursive formulation for an MDP with constraints that may produce suboptimal solutions. It is not applicable in our case because their approach requires a nonnegative reward function.

It should be noted that the aforementioned approaches based on the variability of the return are not suited for problems like the grid world problem discussed in section 6, or the tank control task in section 7 where risk is related to the parameters (variables) of the state description. For example, in the grid world problem, *all* policies have the same worst case outcome. In regard to approaches based on the variance, we found that a policy leading to the error states as fast as possible does not have a higher variance than one that reaches the goal states as fast as possible. A policy with a small variance can therefore have a large risk (with respect to the probability of entering an error state), which means that we address a different class of control problems. We underpin this claim in section 8.1.3.

Fulkerson, Littman, and Keim (1998) sketch an approach in the framework of probabilistic planning that is similar to ours although based on the complementary notion of **safety**. Fulkerson et al. define safety as the probability of reaching a goal state (see also the BURIDAN system of Kushmerick, Hanks, & Weld, 1994). Fulkerson et al. discuss the problem of finding a plan with minimum cost subject to a constraint on the safety (see also Blythe, 1999). For an episodic MDP with goal states, the safety is 1 minus risk. For continuing tasks or if there are absorbing states that are neither goal nor error states, the safety may correspond to a smaller value. Fulkerson et al. (1998) manipulate (scale) the (uniform) step reward of the undiscounted cost model in order to enforce the agent to reach the goal more quickly (see also Koenig & Simmons, 1994). In contrast, we also consider discounted MDPs, and neither require the existence of goal states. Although we do not change the original reward function, our algorithm in section 5 can be seen as a systematic approach for dealing with the idea of Fulkerson et al. that consists in modification of the relative importance of the original objective (reaching the goal) and the safety. In contrast to the aforementioned approaches belonging to the field of probabilistic planning, which operate on an previously known finite MDP, we have designed an online learning algorithm that uses simulated or actual experiences with the process. By the use of neural network techniques the algorithm can also be applied to continuous-state processes.

Dolgov and Durfee (2004) describe an approach that computes policies that have a constrained probability for violating given resource constraints. Their notion of risk is similar to that described by Geibel (2001). The algorithm given by Dolgov and Durfee (2004) computes suboptimal policies using linear programming techniques that require a previously known model and, in contrast to our approach, cannot be easily extended to continuous state spaces. Dolgov and Durfee included a discussion on DP approaches for constrained MDPs (e.g. Altman, 1999) that also do not generalize to continuous state spaces (as in the tank control task) and require a known model. The algorithm described by Feinberg and Shwartz (1999) for constrained problems with two criteria is not applicable in our case, because it requires both discount factors to be strictly smaller than 1, and because it is limited to finite MDPs.

"Downside risk" is a common notion in finance that refers to the likelihood of a security or other investment declining in price, or the amount of loss that could result from such potential decline. The scientific literature on downside risk (e.g. Bawas, 1975; Fishburn, 1977; Markowitz, 1959; Roy, 1952) investigates risk-measures that particularly consider the case in which a return lower than its mean value, or below some target value is encountered. In contrast, our notion of risk is not coupled with the return $R$, but with the fact that a state $x$ is an error state, for example, because some parameters describing the state lie outside their permissible ranges, or because the state lies inside an obstacle which may occur in robotics applications.

## 4. Risk

To define our notion of risk more precisely, we consider a set

$$\Phi \subseteq X \tag{4}$$

of **error states**. Error states are terminal states. This means that the control of the agent ends when it reaches a state in $\Phi$. We allow an additional set of non-error terminal states $\Gamma$ with $\Gamma \cap \Phi = \emptyset$.

Now, we define the risk of $x$ with respect to $\pi$ as the probability that the state sequence $(x_i)_{i \geq 0}$ with $x_0 = x$, which is generated by executing policy $\pi$, terminates in an error state $x' \in \Phi$.

**Definition 4.1 (Risk)** *Let $\pi$ be a policy, and let $x$ be some state. The risk is defined as*

$$\rho^\pi(x) = \mathbf{P}\Big( \exists i\, x_i \in \Phi \,|\, x_0 = x \Big). \tag{5}$$

By definition, $\rho^\pi(x) = 1$ holds if $x \in \Phi$. If $x \in \Gamma$, then $\rho^\pi(x) = 0$ because of $\Phi \cap \Gamma = \emptyset$. For states $\notin \Phi \cup \Gamma$, the risk depends on the action choices of the policy $\pi$.

In the following subsection, we will consider the computation of minimum-risk policies analogous to the computation of maximum-value policies.

## 4.1 Risk Minimization

The risk $\rho^\pi$ can be considered a value function defined for a *cost* signal $\bar{r}$. To see this, we augment the state space of the MDP with an additional absorbing state $\eta$ to which the agent is transfered after reaching a state from $\Phi \cup \Gamma$. The state $\eta$ is introduced for technical reasons.

If the agent reaches $\eta$ from a state in $\Gamma$, both the reward signals $r$ and $\bar{r}$ become zero. We set $r = 0$ and $\bar{r} = 1$, if the agent reaches $\eta$ from an error state. Then the states in $\Phi \cup \Gamma$ are no longer absorbing states. The new cost function $\bar{r}$ is defined by

$$\bar{r}_{x,u}(x') = \begin{cases} 1 & \text{if } x \in \Phi \text{ and } x' = \eta \\ 0 & \text{else.} \end{cases} \tag{6}$$

With this construction of the cost function $\bar{r}$, an episode of states, actions and costs starting at some initial state $x$ contains *exactly* once the cost of $\bar{r} = 1$ if an error state occurs in it. If the process does not enter an error state, the sequence of $\bar{r}$-costs contains zeros only. Therefore, the probability defining the risk can be expressed as the expectation of a cumulative return.

**Proposition 4.1** *It holds*

$$\rho^\pi(x) = \mathbf{E}\left[\sum_{i=0}^{\infty} \bar{\gamma}^i \bar{r}_i \,\Big|\, x_0 = x\right] \tag{7}$$

*with the "discount" factor $\bar{\gamma} = 1$.*

**Proof:** $\bar{r}_0, \bar{r}_1, \ldots$ is the probabilistic sequence of the costs related to the risk. As stated above, it holds that $\bar{R} =_{def} \sum_{i=0}^{\infty} \bar{\gamma}^i \bar{r}_i = 1$ if the trajectory leads to an error state; otherwise $\sum_{i=0}^{\infty} \bar{\gamma}^i \bar{r}_i = 0$. This means that the return $\bar{R}$ is a Bernoulli random variable, and the probability $q$ of $\bar{R} = 1$ corresponds to the risk of $x$ with respect to $\pi$. For a Bernoulli random variable it holds that $\mathbf{E}\bar{R} = q$ (see e.g. Ross, 2000). Notice that the introduction of $\eta$ together with the fact that $\bar{r} = 1$ occurs during the transition from an error state to $\eta$, and not when *entering* the respective error state, ensures the correct value of $\mathbf{E}\left[\sum_{i=0}^{\infty} \bar{\gamma}^i \bar{r}_i \,\Big|\, x_0 = x\right]$ also for error states $x$. q.e.d.

Similar to the Q-function we define the state/action risk as

$$\begin{align}
\bar{Q}^\pi(x,u) &= \mathbf{E}\left[\bar{r}_0 + \bar{\gamma}\rho^\pi(x_1) \mid x_0 = x, u_0 = u\right] \tag{8}\\
&= \sum_{x'} p_{x,u}(x')\left(\bar{r}_{x,u}(x') + \bar{\gamma}\rho^\pi(x')\right). \tag{9}
\end{align}$$

Minimum-risk policies can be obtained with a variant of the Q-learning algorithm (Geibel, 2001).

## 4.2 Maximized Value, Constrained Risk

In general, one is not interested in policies with minimum risk. Instead, we want to provide a parameter $\omega$ that specifies the risk we are willing to accept. Let $X' \subseteq X$ be the set of states we are interested in, e.g. $X' = X - (\Phi \cup \{\eta\})$ or $X' = \{x_0\}$ for a distinguished starting state $x_0$. For a state $x \in X'$, let $p_x$ be the probability for selecting it as a starting state. The value of

$$\mathcal{V}^\pi =_{def} \sum_{x \in X'} p_x V^\pi(x) \qquad (10)$$

corresponds to the performance on the states in $X'$. We consider the constrained problem

$$\max_\pi \mathcal{V}^\pi \qquad (11)$$

subject to

$$\text{for all} \ \ x \in X' : \ \rho^\pi(x) \leq \omega \,. \qquad (12)$$

A policy that fulfills (12) will be called *feasible*. Depending on $\omega$, the set of feasible policies may be empty. Optimal policies generally depend on the starting state, and are non-stationary and randomized (Feinberg & Shwartz, 1999; Gabor et al., 1998; Geibel, 2001). If we restrict the considered policy class to stationary deterministic policies, the constrained problem is generally only well defined if $X'$ is a singleton, because there need not be a stationary deterministic policy being optimal for *all* states in $X'$. Feinberg and Shwartz (1999) have shown for the case of two unequal discount factors smaller than 1 that there exist optimal policies that are randomized Markovian until some time step $n$ (i.e. they do not depend on the history, but may be non-stationary and randomized), and are stationary deterministic (particularly Markovian) from time step $n$ onwards. Feinberg and Shwartz (1999) give a DP algorithm for this case (cp. Feinberg & Shwartz, 1994). This cannot be applied in our case because $\bar{\gamma} = 1$, and also because it does not generalize to continuous state spaces. In the case of *equal* discount factors, it is shown by Feinberg and Shwartz (1996) that (for a fixed starting state) there also exist optimal stationary randomized policies that in the case of one constraint consider at most one action more than a stationary deterministic policy, i.e. there is at most one state where the policy chooses randomly between two actions.

## 5. The Learning Algorithm

For reasons of efficiency and predictability of the agent's behavior and because of what has been said at the end of the last section, we will restrict our consideration to stationary deterministic policies. In the following we present a heuristic algorithm that aims at computing a good policy. We assume that the reader is familiar with Watkin's Q-learning algorithm (Watkins, 1989; Watkins & Dayan, 1992; Sutton & Barto, 1998).

### 5.1 Weighting Risk and Value

We define a new (third) value function $V_\xi^\pi$ and a state/action value function $Q_\xi^\pi$ that is the weighted sum of the risk and the value with

$$V_\xi^\pi(x) = \xi V^\pi(x) - \rho^\pi(x) \qquad (13)$$
$$Q_\xi^\pi(x,u) = \xi Q^\pi(x,u) - \bar{Q}^\pi(x,u) \,. \qquad (14)$$

The parameter $\xi \geq 0$ determines the influence of the $V^\pi$-values ($Q^\pi$-values) compared to the $\rho^\pi$-values ($\bar{Q}^\pi$-values). For $\xi = 0$, $V_\xi^\pi$ corresponds to the negative of $\rho^\pi$. This means that the maximization of $V_0^\pi$ will lead to a minimization of $\rho^\pi$. For $\xi \to \infty$, the maximization of $V_\xi^\pi$ leads to a lexicographically optimal policy for the unconstrained, unweighted 2-criteria problem. If one compares the performance of two policies lexicographically, the criteria are ordered. For large values of $\xi$, the original value function multiplied by $\xi$ dominates the weighted criterion.

The weight is successively adapted starting with $\xi = 0$, see section 5.3. Before adaptation of $\xi$, we will discuss how learning for a fixed $\xi$ proceeds.

## 5.2 Learning for a fixed $\xi$

For a *fixed* value of $\xi$, the learning algorithm computes an optimal policy $\pi_\xi^*$ using an algorithm that resembles Q-Learning and is also based on the ARTDP approach by Gabor et al. (1998).

During learning, the agent has estimates $Q^t$, $\bar{Q}^t$ for time $t \geq 0$, and thus an estimate $Q_\xi^t$ for the performance of its current *greedy* policy, which is the policy that selects the best action with respect to the current estimate $Q_\xi^t$. These values are updated using example state transitions: let $x$ be the current state, $u$ the chosen action, and $x'$ the observed successor state. The reward and the risk signal of this example state transition are given by $r$ and $\bar{r}$ respectively. In $x'$, the greedy action is defined in the following manner: an action $u$ is preferable to $u'$ if $Q_\xi^t(x', u) > Q_\xi^t(x', u')$ holds. If the equality holds, the action with the higher $Q^t$-value is preferred. We write $u \succeq u'$, if $u$ is preferable to $u'$.

Let $u^*$ be the greedy action in $x'$ with respect to the ordering $\succeq$. Then the agent's estimates are updated according to

$$Q^{t+1}(x, u) = (1 - \alpha_t)Q^t(x, u) + \alpha_t(r + \gamma Q^t(x', u^*)) \tag{15}$$

$$\bar{Q}^{t+1}(x, u) = (1 - \alpha_t)\bar{Q}^t(x, u) + \alpha_t(\bar{r} + \bar{\gamma}\bar{Q}^t(x', u^*)) \tag{16}$$

$$Q_\xi^{t+1}(x, u) = \xi Q^{t+1}(x, u) - \bar{Q}^{t+1}(x, u) \tag{17}$$

Every time a new $\xi$ is chosen, the learning rate $\alpha_t$ is set to 1. Afterwards $\alpha_t$ decreases over time (cp. Sutton & Barto, 1998).

For a fixed $\xi$, the algorithm aims at computing a good stationary deterministic policy $\pi_\xi^*$ for the weighted formulation that is feasible for the original constrained problem. Existence of an optimal stationary deterministic policy for the weighted problem and convergence of the learning algorithm can be guaranteed if both criteria have the same discount factor, i.e. $\gamma = \bar{\gamma}$, even when $\bar{\gamma} < 1$. In the case $\gamma = \bar{\gamma}$, $Q_\xi$ forms a standard criterion function with rewards $\xi r - \bar{r}$. Because we consider the risk as the second criterion function, $\gamma = \bar{\gamma}$ implies that $\gamma = \bar{\gamma} = 1$. To ensure convergence in this case it is also required that either (a) there exists at least one proper policy (defined as a policy that reaches an absorbing state with probability one), and improper policies yield infinite costs (see Tsitsiklis, 1994), or (b), all policies are proper. This is the case in our application example. We conjecture that in the case $\gamma < \bar{\gamma}$ convergence to a possibly suboptimal policy can be guaranteed if the MDP forms a directed acyclic graph (DAG). In other cases oscillations and non-convergence may occur, because optimal policies for the weighted problem are generally not found in the considered policy class of stationary deterministic policies (as for the constrained problem).

### 5.3 Adaptation of $\xi$

When learning starts, the agent chooses $\xi = 0$ and performs learning steps that will lead, after some time, to an approximated minimum-risk policy $\pi_0^*$. This policy allows the agent to determine if the constrained problem is feasible.

Afterwards the value of $\xi$ is increased step by step until the risk in a state in $X'$ becomes larger than $\omega$. Increasing $\xi$ by some $\epsilon$ increases the influence of the $Q$-values compared to the $\bar{Q}$-values. This may cause the agent to select actions that result in a higher value, but perhaps also in a higher risk. After increasing $\xi$, the agent again performs learning steps until the greedy policy is sufficiently stable. This is aimed at producing an optimal deterministic policy $\pi_\xi^*$. The computed $Q$- and $\bar{Q}$-values for the old $\xi$ (i.e. estimates for $Q^{\pi_\xi^*}$ and $\bar{Q}^{\pi_\xi^*}$) are used as the initialization for computing $\pi_{\xi+\epsilon}^*$.

The aim of increasing $\xi$ is to give the value function $V$ the maximum influence possible. This means that the value of $\xi$ is to be maximized, and needs not be chosen by the user. The adaptation of $\xi$ provides a means for searching the space of feasible policies.

### 5.4 Using a Discounted Risk

In order to prevent oscillations of the algorithm in section 5.2 for the case $\gamma < \bar{\gamma}$, it may be advisable to set $\bar{\gamma} = \gamma$ corresponding to using a **discounted risk** defined as

$$\rho_\gamma^\pi(x) = \mathbf{E}\left[\sum_{i=0}^{\infty} \gamma^i \bar{r}_i \,\Big|\, x_0 = x\right].\tag{18}$$

Because the values of the $\bar{r}_i$ are all positive, it holds $\rho_\gamma^\pi(x) \leq \rho^\pi(x)$ for all states $x$. The discounted risk $\rho_\gamma^\pi(x)$ gives more weight to error states occurring in the near future, depending on the value of $\gamma$.

For a finite MDP and a fixed $\xi$, the convergence of the algorithm to an optimal stationary policy for the weighted formulation can now be guaranteed because $Q_\xi$ (using $\rho_\gamma^\pi(x)$) forms a standard criterion function with rewards $\xi r - \bar{r}$. For terminating the adaptation of $\xi$ in the case that the risk of a state in $X'$ becomes larger than $\omega$, one might still use the original (undiscounted) risk $\rho^\pi(x)$ while learning is done with its discounted version $\rho_\gamma^\pi(x)$, i.e. the learning algorithm has to maintain two risk estimates for every state, which is not a major problem. Notice that in the case of $\gamma = \bar{\gamma}$, the *effect* of considering the weighted criterion $\xi V^\pi - \rho_\gamma^\pi$ corresponds to modifying the *unscaled* original reward function $r$ by adding a negative reward of $-\frac{1}{\xi}$ when the agent enters an error state: the set of optimal stationary deterministic policies is equal in both cases (where the added absorbing state $\eta$ with its single dummy action can be neglected).

In section 6, experiments for the case of $\gamma < 1 = \bar{\gamma}$, $X' = X - (\Phi \cup \{\eta\})$, and a finite state space can be found. In the sections 7 and 8 we will consider an application example with infinite state space, $X' = \{x_0\}$, and $\gamma = \bar{\gamma} = 1$.

## 6. Grid World Experiment

In the following we will study the behaviour of the learning algorithm for a finite MDP with a discounted criterion. In contrast to the continuous-state case discussed in the next section, no function approximation with neural networks is needed because both the value function and the risk can be stored in a table. For the grid world, we have chosen $\gamma < 1 = \bar{\gamma}$, $X' = X - \Phi$, and a state graph that is not a DAG. This implies that there is no stationary policy which is optimal for every state in $X'$. Although oscillations can therefore be expected, we have found that the algorithm stabilizes at a feasible policy because the learning rate $\alpha_t$ tends to zero. We also investigated the use of the discounted risk that prevents an oscillatory behaviour.

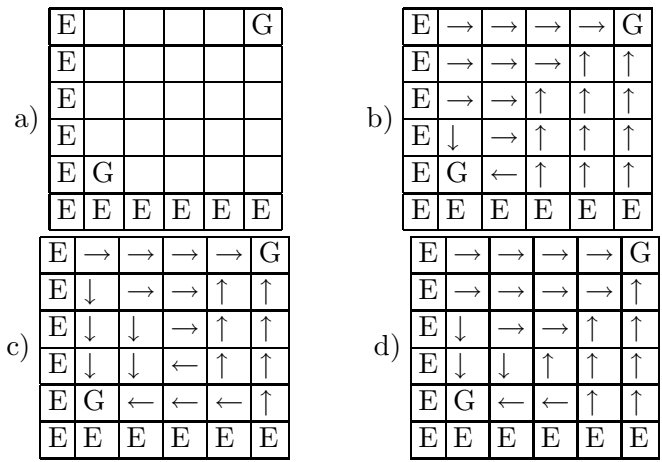

Figure 1: a) An example grid world, $x$ : horizontal, $y$ : vertical. For further explanation see text. b) Minimum risk policy ($\xi = 0$) with 11 unsafe states. c) Maximum value policy ($\xi = 4.0$) with 13 unsafe states. d) Result of algorithm: policy for $\xi = 0.64$ with 11 unsafe states.

We consider the $6 \times 6$ grid world that is depicted in Figure 1(a). An empty field denotes some state, $E$s denote error states, and the two $G$s denote two goal states. We describe states as pairs $(x, y)$ where $x, y \in \{1, 2, 3, 4, 5, 6\}$. I.e. $\Gamma = \{(2, 2), (6, 6)\}$, $\Phi = \{(1, 1), (1, 2), (1, 3), (1, 4), (1, 5), (1, 6), (2, 1), (3, 1), (4, 1), (5, 1), (6, 1)\}$. The additional absorbing state $\eta$ is not depicted.

We have chosen the error states as if the lower, i.e. extremal values of $x$ and $y$ were dangerous. One of the goal states is placed next to the error states, the other in a safer part of the state space.

The agent has the actions $U = \{\rightarrow, \leftarrow, \uparrow, \downarrow\}$. An action $u \in U$ takes the agent in the denoted direction if possible. With a probability of 0.21, the agent is not transported to the desired direction but to one of the three remaining directions.

The agent receives a reward of 1 if it enters a goal state. The agent receives a reward of 0 in every other case. It should be noted that there is no explicit punishment for entering an error state, but there is an implicit one: if the agent enters an error state, then the current episode ends. This means that the agent will never receive a positive reward after it has reached an error state. Therefore, it will try to reach one of the goal states, and because $\gamma < 1$, it will try to do this as fast as possible.

We have chosen $X' = X - (\Phi \cup \{\eta\})$, $\gamma = 0.9$, and equal probabilities $p_x$ for all states. Although the convergence of the algorithm cannot be guaranteed in this case, the experimental results show that the algorithm yields a feasible policy.

We have selected $\omega = 0.13$. In order to illustrate the behaviour of the algorithm we have also computed the minimum-risk and maximum-value policy. Figure 1(b) shows the minimum risk policy. Though the reward function $r$ defined above plays no role for the minimum risk policy, the agent tries to reach one of the two goal states. This is so because from a goal state the probability of reaching an error state is 0. Clearly, with respect to the value function $V$, the policy in Figure 1(b) is not optimal: e.g. from state $(3, 3)$ the agent tries to reach the more distant goal, which causes higher discounting of the goal reward. The minimum risk policy in Figure 1(b) has 25 *safe* states, defined as states for which the risk is below $\omega$. The minimum risk policy has an estimated mean value of $\mathcal{V}^\pi = 0.442$.

In Figure 1(c) the maximum-value policy is shown. The maximum-value policy that optimizes the value without considering the risk has an estimated value of $\mathcal{V}^\pi = 0.46$. Thus, it performs better than the minimum-risk policy in Figure 1(b), but the risk in $(5, 2)$ and $(2, 5)$ has become greater than $\omega$. Our algorithm starts with $\xi = 0$ and computes the minimum-risk policy in Figure 1(b). $\xi$ is increased step by step until the risk for a state changes from a value lower than $\omega$ to a value $> \omega$. Our algorithm stops at $\xi = 0.64$. The policy computed is shown in Figure 1(d). Obviously, it lies "in between" the minimum risk policy in Figure 1(b) and the maximum-value policy in Figure 1(c).

We also applied the algorithm with the discounted version of the risk, $\rho_\gamma^\pi$, to the grid world problem. The discounted risk was used for learning, whereas the original risk, $\rho^\pi$, was used for selecting the best weight $\xi$. For the parameters described above, the modified algorithm also produced the policy depicted in figure 1(d). Seemingly, in the grid world example, oscillations do not present a major problem.

For the tank control task described in the next section, it holds that $\rho_\gamma^\pi = \rho^\pi$ because $\gamma = \bar{\gamma}$.

## 7. Stochastic Optimal Control with Chance Constraints

In this section, we consider the solution of a stochastic optimal control problem with chance constraints (Li et al., 2002) by applying our risk-sensitive learning method.

### 7.1 Description of the Control Problem

In the following, we consider the plant depicted in Figure 2. The task is to control the outflow of the tank that lies upstream of a distillation column in order to fulfill several objectives that are described below. The purpose of the distillation column is the separation of two substances 1 and 2. We consider a finite number of time steps $0, \ldots, N$. The outflow of the tank, i.e. the feedstream of the distillation column, is characterized by a flowrate $F(t)$ which is controlled by the agent, and the substance concentrations $c_1(t)$ and $c_2(t)$ (for $0 \le t \le N$).

The purpose of the control to be designed is to keep the outflow rate $F(t)$ near a specified optimal flow rate $\mathrm{F_{spec}}$ in order to guarantee optimal operation of the distillation column.

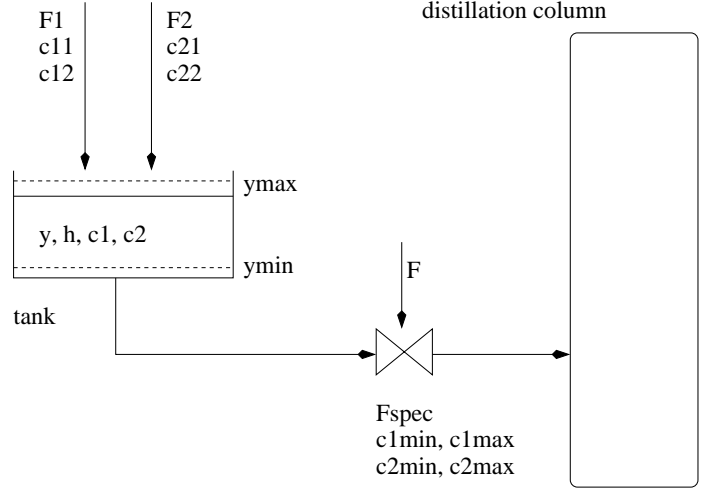

Figure 2: The plant. See text for description.

Using a quadratic objective function, this goal is specified by

$$\min_{F(0),...,F(N-1)} \sum_{t=0}^{N-1} (F(t) - F_{\text{spec}})^2,$$  (19)

where the values obey

$$\text{for } 0 \le t \le N-1 : F_{\min} \le F(t) \le F_{\max}.$$  (20)

The tank is characterized by its tank level $y(t)$ and its holdup $h(t)$, where $y = A^{-1}h$ with some constant $A$ for the footprint of the tank. The tank level $y(t)$ and the concentrations $c_1(t)$ and $c_2(t)$ depend on the two stochastic inflow streams characterized by the flowrates $F_1(t)$ and $F_2(t)$, and the inflow concentrations $c_{1,j}(t)$ and $c_{2,j}(t)$ for substances $j \in \{1, 2\}$. The linear dynamics of the tank level is given by

$$y(t+1) = y(t) + A^{-1}\Delta t \cdot \Big( \sum_{j=1,2} F_j(t) - F(t) \Big).$$  (21)

The dynamics of the concentrations is given by

$$\text{for } i = 1, 2 : c_i(t+1) = c_i(t) + \frac{A^{-1}\Delta t}{y(t)} \big( \sum_{j=1,2} F_j(t)(c_{j,i}(t) - c_i(t)) \big)$$  (22)

The initial state of the system is characterized by

$$y(0) = y_0, c_1(0) = c_1^0, c_2(0) = c_2^0.$$  (23)

The tank level is required to fulfill the constraint $y_{\min} \le y(t) \le y_{\max}$. The concentrations inside the tank correspond to the concentrations of the outflow. The substance concentrations $c_1(t)$ and $c_2(t)$ are required to remain in the intervals $[c_{1,\min}, c_{1,\max}]$ and $[c_{2,\min}, c_{2,\max}]$, respectively. We assume that the inflows $F_i(t)$ and inflow concentrations $c_{i,j}(t)$ are random, and that they are governed by a probability distribution. Li et al. (2002) assume a multivariate Gaussian distribution. Because of the randomness of the variables, the tank level and the feedstream concentrations may violate the given constraints. We therefore formulate the stochastic constraint

$$\mathbf{P}\Big( y_{\min} \leq y(t) \leq y_{\max}, c_{i,\min} \leq c_i(t) \leq c_{i,\max}, 1 \leq t \leq N, i = 1, 2 \Big) \geq p \qquad (24)$$

The expression in (24) is called a (joint) **chance constraint**, and $1 - p$ corresponds to the permissible probability of constraint violation. The value of $p$ is given by the user.

The stochastic optimization problem `SOP-YC` is defined by the quadratic objective function (19) describing the sum of the quadratic differences of the outflow rates and $F_{spec}$, the linear dynamics of the tank level in (21), the nonlinear dynamics of the concentrations in (22), the initial state given in (23), and the chance constraint in (24).

Li et al. describe a simpler problem `SOP-Y` where the concentrations are not considered; see Figure 3. For `SOP-Y` we use the cumulative inflow $F_\Sigma = F_1 + F_2$ in the description of the tank level dynamics, see (27). `SOP-Y` describes the dynamics of a *linear* system. Li et al. solve `SOP-Y` by relaxing it to a nonlinear program that is solved by sequential quadratic programming. This relaxation is possible because `SOP-Y` is a linear system, and a multivariate Gaussian distribution is assumed. Solving of nonlinear systems like `SOP-YC` and non-Gaussian distributions is difficult (e.g. Wendt, Li, & Wozny, 2002), but can be achieved with our RL approach.

---

$$\min_{F(0),\dots,F(N-1)} \qquad \sum_{t=0}^{N-1} (F(t) - F_{spec})^2 \qquad (25)$$

$$\text{subject to}$$

$$\text{for } 0 \leq t \leq N - 1: \qquad F_{\min} \leq F(t) \leq F_{\max} \qquad (26)$$

$$y(t+1) \quad = \quad y(t) + A^{-1} \Delta t \cdot \Big( F_\Sigma(t) - F(t) \Big) \qquad (27)$$

$$y(0) = y_0 \qquad (28)$$

$$\mathbf{P}\Big( y_{\min} \leq y(t) \leq y_{\max}, 1 \leq t \leq N \Big) \geq p \qquad (29)$$

---

Figure 3: The problem `SOP-Y`.

Note that the control $F(t)$ in the optimization problems only depends on the time step $t$. This means that the solutions of `SOP-YC` and `SOP-Y` yield open loop controls. Because of the dependence on the initial condition in (23), a moving horizon approach can be taken to design a closed loop control. We will not discuss this issue, as it goes beyond the scope of the paper.

## 7.2 Formulation as a Reinforcement Learning Problem

Using RL instead of an analytical approach has the advantage that the probability distribution doesn't have to be Gaussian and it can be unknown. The state equations also need not be known, and they can be nonlinear. But the learning agent must have access to simulated or empirical data, i.e. samples of at least some of the random variables.

Independent of the chosen state representation, the immediate reward is defined by

$$r_{x,u}(x') = -(u - \text{F}_{\text{spec}})^2, \tag{30}$$

where $u$ is a chosen action – the minus is required because the RL value function is to be *maximized*. The reward signal only depends on the action chosen, not on the current and the successor state.

In this work we only consider finite (discretized) action sets, although our approach can also be extended to continuous action sets, e.g. by using an actor-critic method (Sutton & Barto, 1998). In the following, we assume that the interval $[F_{\text{min}}, F_{\text{max}}]$ is discretized in an appropriate manner.

The process reaches an **error state** if one of the constraints in (24) (or in (29), respectively) is violated. The process is then artificially terminated by transferring the agent to the additional absorbing state $\eta$ giving a risk signal of $\bar{r} = 1$. The $V^*$-value of error states is set to zero, because the controller could choose the action $\text{F}_{\text{spec}}$ after the first constraint violation, as subsequent constraint violations do not make things worse with respect to the chance constraints (24) and (29), respectively.

## 7.3 Definition of the State Space

In the following we consider the design of appropriate state spaces that result either in an open loop control (OLC) or a closed loop control (CLC).

### 7.3.1 OPEN LOOP CONTROL

We note that SOP-YC and SOP-Y are time-dependent finite horizon problems where the control $F(x_t) = F(t)$ depends on $t$ only. This means that there is no state feedback and the resulting controller is open-looped. With respect to the state definition $x_t = (t)$, the Markov property defined in section 2 clearly holds for the probabilities and rewards defining $V^\pi$. But the Markov property does not hold for the rewards defining $\rho^\pi$. Using $x_t = (t)$ implies that the agent has no information about the state of the process. By including information about the history in the form of its past action, the agent gets an "idea" about the current state of the process. Therefore, the inclusion of history information changes the probability for $\bar{r} = 1$, and the Markov property is violated. Including the past actions in the state description ensures the Markov property for $\bar{r}$. The Markov property is therefore recovered by considering the augmented state definition

$$x_t = (t, u_{t-1}, \ldots, u_0), \tag{31}$$

with past actions $(u_{t-1}, \ldots, u_0)$. The first action $u_0$ depends on the fixed initial tank level $y_0$ and the fixed initial concentrations only. The second action depends on the first action, i.e. also on the initial tank level and the initial concentrations and so on. Therefore, learning with states (31) results in an open loop control, as in the original problems `SOP-YC` and `SOP-Y`.

It should be noted that for an MDP, the risk does not depend on past actions, but on future actions only. For the choice $x_t = (t)$, there is hidden state information, and we do *not* have an MDP because the Markov property is violated. Therefore the probability of entering an error state *conditioned on the time step*, i.e. $P(\bar{r}_0 = 1|t)$, changes if it is additionally conditioned on the past actions yielding the value $P(\bar{r}_0 = 1|t, u_{t-1}, \ldots, u_0)$ (corresponding to an agent that remembers its past actions). For example, if the agent remembers that in the past time steps of the current learning episode it has always used action $F = 0$ corresponding to a zero outflow, it can conclude that there is an increased probability that the tank level exceeds $y_{\max}$, i.e. it can have knowledge of an increased risk. If, on the other hand, it does not remember its past actions, it cannot know of an increased risk because it only knows the index of the current time step, which carries less information about the current state.

It is well-known that the Markov property can generally be recovered by including the complete state history into the state description. For $x_t = (t)$, the state history contains the past time indices, actions and $\bar{r}$-costs. For the tank control task, the action history is the relevant part of the state history because all previous $\bar{r}$-costs are necessarily zero, and the indices of the past time steps are already given with the actual time $t$ that is known to the agent. Therefore, the past rewards and the indices of the past time steps need not be included into the expanded state. Although still not the complete state information is known to the agent, knowledge of past actions suffices to recover the Markov property.

With respect to the state choice (31) and the reward signal (30), the expectation from the definition of the value function is not needed, cp. eq. (2). This means that

$$V^\pi(x) = \mathbf{E}\left[R \,|\, x_0 = x\right] = -\sum_{t=0}^{N-1} (F(t) - \mathrm{F_{spec}})^2$$

holds, i.e. there is a direct correspondence between the value function and the objective function of `SOP-YC` and `SOP-Y`.

### 7.3.2 Closed Loop Control

We will now define an alternative state space, where the expectation *is* needed. We have decided to use the state definition

$$x_t = (t, y(t), c_1(t), c_2(t)) \tag{32}$$

for the problem `SOP-YC` and

$$x_t = (t, y(t)) \tag{33}$$

for the simpler problem `SOP-Y`. The result of learning is a state and time-dependent closed loop controller, which can achieve a better regulation behavior than the open loop controller, because it reacts on the actual tank level and concentrations, whereas an open loop control does not. If the agent has access to the inflow rates or concentrations, they too can be included in the state vector, yielding improved performance of the controller.

Table 1: Parameter settings

| Parameter | Value | Explanation |
|---|---|---|
| $N$ | 16 | number of time steps |
| $y_0$ | 0.4 | initial tank level |
| $[y_{\min}, y_{\max}]$ | $[0.25, 0.75]$ | admissible interval for tank level |
| $A^{-1}\Delta t$ | 0.1 | constant, see (22) |
| $F_{\text{spec}}$ | 0.8 | optimal action value |
| $[F_{\min}, F_{\max}]$ | $[0.55, 1.05]$ | interval for actions, 21 discrete values |
| only RL-YC-CLC: | | |
| $c_1^0$ | 0.2 | initial concentration subst. 1 |
| $c_2^0$ | 0.8 | initial concentration subst. 2 |
| $[c_{1,\min}, c_{1,\max}]$ | $[0.1, 0.4]$ | interval for concentration 1 |
| $[c_{2,\min}, c_{2,\max}]$ | $[0.6, 0.9]$ | interval for concentration 2 |

## 7.4 The RL Problems

With the above definitions, the optimization problem is defined via (11) and (12) with $\omega = 1 - p$ (see (24) and (29)). The set $X'$ (see (10) and (12)) is defined to contain the unique starting state, i.e $X' = \{x_0\}$. In our experiments we consider the following instantiations of the RL problem:

- RL-Y-CLC Reduced problem SOP-Y using states $x_t = (t, y(t))$, with $x_0 = (0, y_0)$ resulting in a closed loop controller (CLC).

- RL-Y-OLC Open loop controller (OLC) for reduced problem SOP-Y. The state space is defined by the action history and time, see eq. (31). The starting state is $x_0 = (0)$.

- RL-YC-CLC Closed loop controller for full problem SOP-YC using states $x_t = (t, y(t), c_1(t), c_2(t))$ with $x_0 = (0, y_0, c_1^0, c_1^0)$.

Solving the problem RL-Y-OLC yields an action vector. The problems RL-YC-CLC and RL-Y-CLC result in state dependent controllers. We do not present results for the fourth natural problem RL-YC-OLC, because they offer no additional insights.

For interpolation between states we used $2 \times 16$ multilayer perceptrons (MLPs, e.g. Bishop, 1995) in the case of RL-Y-OLC because of the extremely large state space (15 dimensions for $t = N - 1$). We used radial basis function (RBF) networks in the case of RL-YC-CLC and RL-Y-CLC, because they produced faster, more stable and robust results compared to MLPs.

For training the respective networks, we used the "direct method" that corresponds to performing one gradient descent step for the current state-action pair with the new estimate as the target value (see e.g. Baird, 1995). The new estimate for $Q^{\pi_\xi^*}$ is given by $r + \gamma Q^t(x', u^*)$, and for $\bar{Q}^{\pi_\xi^*}$ by $\bar{r} + \bar{\gamma}\bar{Q}^t(x', u^*)$ (compare the right sides of the update equations (15)-(17)).

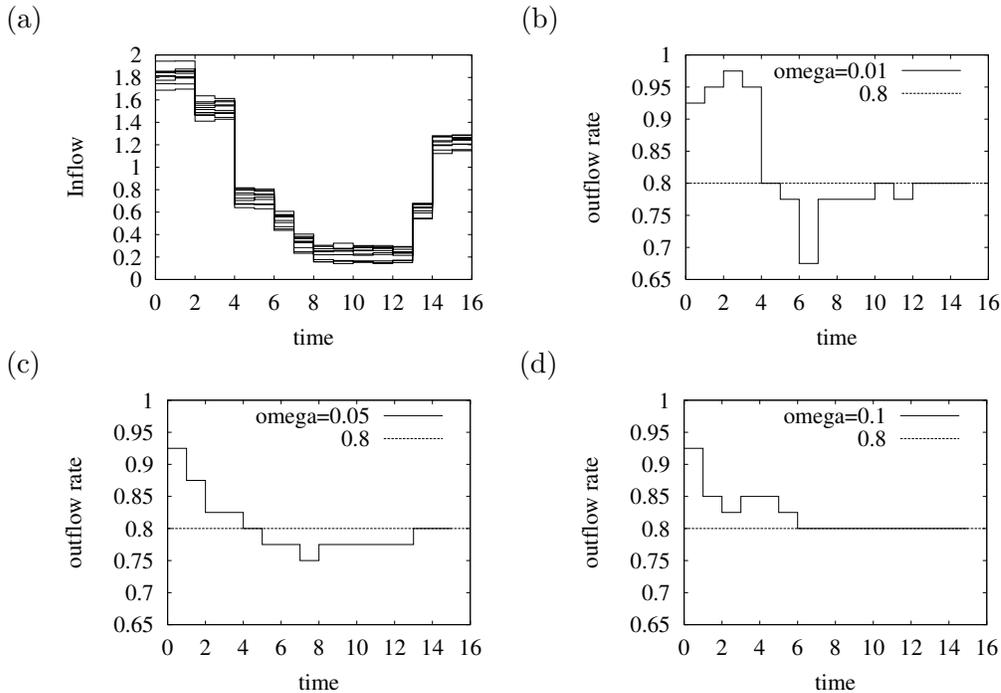

Figure 4: `RL-Y-CLC`: (a) The inflow rates $F_\Sigma(t)$ for 10 runs. (b), (c), (d) Example runs of policies for $\omega = 0.01, 0.05, 0.10$ (i.e. $p = 0.99, 0.95, 0.90$). It holds $F_{spec} = 0.8$.

## 8. Experiments

In this section, we examine the experimental results obtained for the tank control task ($\gamma = \bar{\gamma} = 1$). In section 8.1 we discuss the linear case and compare the results to those of Li et al. (2002). For the linear case, we consider the closed loop controller obtained by solving `RL-Y-CLC` (sect. 8.1.1) and the open loop controller related to the RL problem `RL-Y-OLC` (sect. 8.1.2). For the closed loop controller, we discuss the problem of non-zero covariances between variables of different time steps. The nonlinear case is discussed in section 8.2.

### 8.1 The Problems `RL-Y-CLC` and `RL-Y-OLC`

We start with the simplified problems, `RL-Y-CLC` and `RL-Y-OLC`, derived from `SOP-Y` that is discussed by Li et al. (2002). In `SOP-Y` the concentrations are not considered, and there is only one inflow rate $F_\Sigma(t) = F_1(t) + F_2(t)$. The parameter settings in Table 1 (first five lines) were taken from Li et al. (2002). The minimum and maximum values for the actions were determined by preliminary experiments.

Li et al. define the inflows $(F_\Sigma(0), \dots, F_\Sigma(15))^T$ as having a Gaussian distribution with the mean vector

$$(1.8, 1.8, 1.5, 1.5, 0.7, 0.7, 0.5, 0.3, 0.2, 0.2, 0.2, 0.2, 0.2, 0.6, 1.2, 1.2)^T . \tag{34}$$

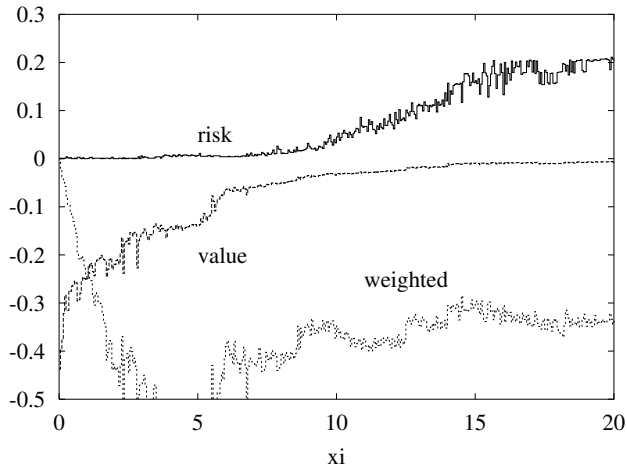

Figure 5: `RL-Y-CLC`: Estimates of the risk $\rho^{\pi_\xi^*}(x_0)$, the value $V^{\pi_\xi^*}(x_0)$, and of $V_\xi^{\pi_\xi^*}(x_0) = \xi V^{\pi_\xi^*}(x_0) - \rho^{\pi_\xi^*}(x_0)$ for different values of $\xi$.

The covariance matrix is given by

$$C = \begin{pmatrix} \sigma_0^2 & \sigma_0\sigma_1 r_{01} & \cdots & \sigma_0\sigma_{N-1} r_{0(N-1)} \\ \sigma_0\sigma_1 r_{01} & \cdots & \cdots & \cdots \\ \cdots & \cdots & \cdots & \cdots \\ \sigma_0\sigma_{N-1} r_{0(N-1)} & \cdots & \cdots & \sigma_{N-1}^2 \end{pmatrix} \tag{35}$$

with $\sigma_i = 0.05$. The correlation of the inflows of time $i$ and $j$ is defined by

$$r_{ij} = r_{ji} = 1 - 0.05(j - i) \tag{36}$$

for $0 \le i \le N - 1, i < j \le N - 1$ (from Li et al., 2002). The inflow rates for ten example runs are depicted in Figure 4(a).

### 8.1.1 THE PROBLEM `RL-Y-CLC` (CONSTRAINTS FOR TANK LEVEL)

We start with the presentation of the results for the problem `RL-Y-CLC`, where the control (i.e. the outflow $F$) depends only on the time $t$ and the tank level. Because $X' = \{x_0\}$ the overall performance of the policy as defined in (10) corresponds to its performance for $x_0$,

$$\mathcal{V}^{\pi_\xi^*} = V^{\pi_\xi^*}(x_0).$$

It holds that $x_0 = (0, y_0)$. $V^{\pi_\xi^*}(x_0)$ is the value with respect to the policy $\pi_\xi^*$ *learned* for the weighted criterion function $V_\xi^\pi$, see also (13). The respective risk is

$$\rho^{\pi_\xi^*}(x_0).$$

In Figure 5 the estimated[2] risk $\rho^{\pi_\xi^*}(x_0)$ and the estimated value $V^{\pi_\xi^*}(x_0)$ are depicted for different values of $\xi$. Both the estimate for the risk $\rho^{\pi_\xi^*}(x_0)$ and that for value $V^{\pi_\xi^*}(x_0)$ increase with $\xi$. Given a fixed value $p$ for the admissible probability of constraint violation, the appropriate $\xi = \xi(p)$ can be obtained as the value for which the risk $\rho^{\pi_\xi^*}(x_0)$ is lower than $\omega = 1 - p$ and has the maximum $V^{\pi_\xi^*}(x_0)$. Due to the variation of the performance (see Fig. 5) we found that this works better than just selecting the maximum $\xi$. The estimate of the weighted criterion $V_\xi^{\pi_\xi^*}(x_0) = \xi V^{\pi_\xi^*}(x_0) - \rho^{\pi_\xi^*}(x_0)$ is also shown in Figure 5.

---

2. All values and policies presented in the following were estimated by the learning algorithm. Note that in order to enhance the readability, we have also denoted the *learned* policy as $\pi_\xi^*$.

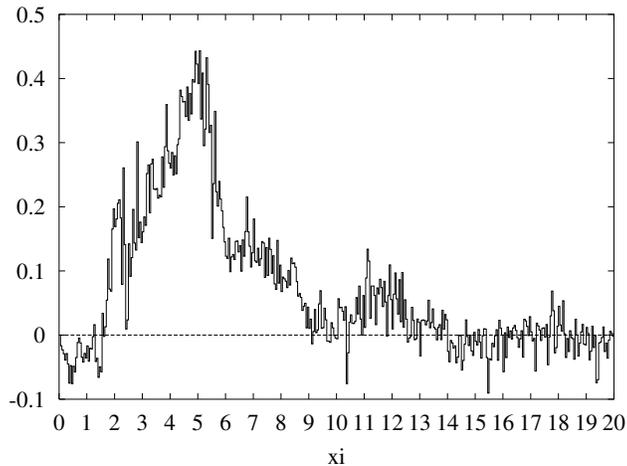

Figure 6: `RL-Y-CLC`: Difference between the weighted criteria. For an explanation see text.

The outflow rate $F$ (control variable) for different values of $\omega$ can be found in Figure 4(b-c). Note that the rates have a certain variance since they depend on the probabilistic tank level. We randomly picked one example run for each value of $\omega$. It is found that the control values $F(t)$ tend to approach $F_{\text{spec}}$ with increasing values of $\omega$ (i.e. decreasing values of $p$).

**Correlations** The definition of the covariance matrix in (35) and (36) reveals a high correlation of the inflow rates in neighboring time steps. In order to better account for this, it is possible to include information on past time steps in the state description at time $t$. Because the level $y$ changes according to the inflow rate $F_\Sigma$, we investigated the inclusion of past values of $y$. If the inflow rates were measured, they too could be included in the state vector. Former rewards need not be included because they depend on the past tank levels, i.e. they represent redundant information.

We have compared the performance of the algorithm for the augmented state space defined by $\tilde{x}_t = (t, y(t), y(t-1), y(t-2))$ (depth 2 history) and the normal state space $x_t = (t, y(t))$ (no history). Fig. 6 shows

$$V^{\tilde{\pi}_\xi^*}\Big(\underbrace{(0, y_0, 0, 0)}_{\tilde{x}_0}\Big) - V^{\pi_\xi^*}\Big(\underbrace{(0, y_0)}_{x_0}\Big),$$

i.e. the difference in the weighted criteria for the starting state with respect to the learned policies $\tilde{\pi}_\xi^*$ (history) and $\pi_\xi^*$ (no history). Note that for the starting state $\tilde{x}_0$, the past values have been defined as 0. The curve in Figure 6 runs mainly above 0. This means that using the augmented state space results in a better performance for many values of $\xi$. Note that for larger values of $\xi$ the original value function overweights the risk so that in both cases the policy that always chooses the outflow $F_{spec}$ is approximated. This means that the difference in performance tends to zero.

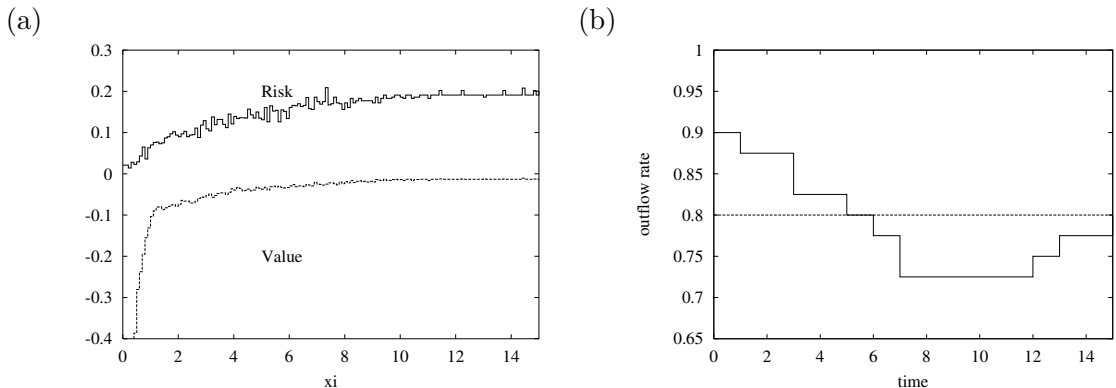

Figure 7: `RL-Y-OLC`: (a) Estimates of the risk $\rho^{\pi_\xi^*}(x_0)$ and the value $V^{\pi_\xi^*}(x_0)$ for increasing values of $\xi$. (b) Learned policy $\pi_\xi^*$ with $\rho^{\pi_\xi^*}(x_0) \approx 0.098$ and $V^{\pi_\xi^*}(x_0) \approx -0.055$

A similar, but not quite pronounced effect can be observed when using a history of length 1 only. In principle, we assume that it is possible to achieve even better performance by including the full history of tank levels in the state description, but there is a trade-off between this objective and the difficulty of network training caused by the number of additional dimensions.

### 8.1.2 `RL-Y-OLC` (History of Control Actions)

The RL problem `RL-Y-OLC` comprises state descriptions consisting of an action history together with the time, see eq. (31). The starting state has an empty history, i.e. $x_0 = (0)$. The result of learning is a time-dependent policy with an implicit dependence on $y_0$. The learned policy $\pi_\xi^*$ is therefore a fixed vector of actions $F(0), \ldots, F(15)$ that forms a feasible, but in general suboptimal solution to the problem `SOP-Y` in Figure 3.

The progression of the risk estimate, i.e. of $\rho^{\pi_\xi^*}(x_0)$, and that for the value, $V^{\pi_\xi^*}(x_0)$, for different values of $\xi$ can be found in Figure 7. The results are not as good as the ones for `RL-Y-CLC` in Figure 5: the estimated minimum risk is 0.021, and the risk $\rho^{\pi_\xi^*}(x_0)$ grows much faster than the `RL-Y-CLC`-risk in Figure 5.

A policy having a risk $\rho^{\pi_\xi^*}(x_0) \approx 0.098$ is depicted in Figure 7(b). In contrast to the policies for `RL-Y-CLC` (see Figure 4(b-c)), the control values do not change in different runs.

### 8.1.3 Comparison

In Table 2, we have compared the performance of the approach of Li et al. with `RL-Y-CLC` and `RL-Y-OLC` for $p = 0.8$ and $p = 0.9$. For both `RL-Y-CLC` and `RL-Y-OLC` we performed 10 learning runs. For the respective learned policy $\pi$, the risk $\rho^\pi(x_0)$ and value $V^\pi(x_0)$ were estimated during 1000 test runs. For `RL-Y-CLC` and `RL-Y-OLC`, the table shows the mean performance averaged over this 10 runs together with the standard deviation in parentheses.

Table 2: Comparison of est. squared deviation to $F_{spec}$ (i.e. $-V^\pi(x_0)$) for results of Li et al. with results for `RL-Y-CLC` and `RL-Y-OLC` for $p = 0.8$ ($\omega = 0.2$) and $p = 0.9$ ($\omega = 0.1$). Smaller values are better.

| approach | $p = 0.8$ | $p = 0.9$ |
|---|---|---|
| Li et al. (2002) | 0.0123 | 0.0484 |
| `RL-Y-CLC` | 0.00758 (0.00190) | 0.02 (0.00484) |
| `RL-Y-OLC` | 0.0104 (0.000302) | 0.0622 (0.0047) |

It is found that, in average, the policy determined for `RL-Y-CLC` performs better than that obtained through the approach of Li et al. (2002) (with respect to the estimated squared deviation to the desired outflow $F_{spec}$, i.e. with respect to $-V^\pi(x_0)$.) The policy obtained for `RL-Y-OLC` performs better for $p = 0.8$ and worse for $p = 0.9$. The maximal achievable probability for holding constraints was 1.0 (sd 0.0) for `RL-Y-CLC`, and and 0.99 (sd 0.0073) for `RL-Y-OLC`. Li et al. report $p = 0.999$ for their approach.

The approach of Neuneier and Mihatsch (1999) considers the *worst-case outcomes of a policy*, i.e. risk is related to the variability of the return. Neuneier and Mihatsch show that the learning algorithm interpolates between risk-neutral and the worst-case criterion and has the same limiting behavior as the exponential utility approach.

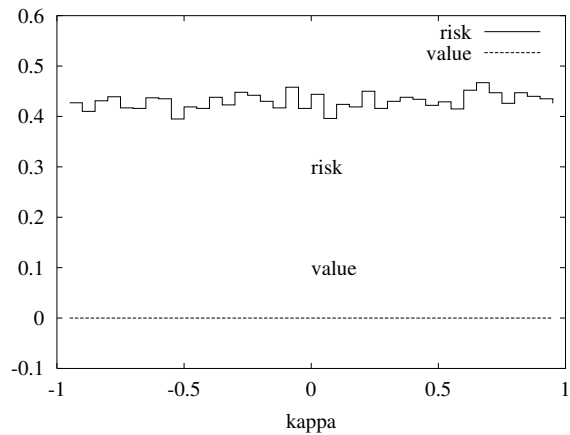

Figure 8: Risk and value for several values of $\kappa$

The learning algorithm of Neuneier and Mihatsch has a parameter $\kappa \in (-1.0, 1.0)$ that allows to switch between risk-averse behavior ($\kappa \to 1$), risk-neutral behavior ($\kappa = 0$), and risk-seeking behavior ($\kappa \to -1$). If the agent is risk-seeking, it prefers policies with a good best-case outcome. Figure 8 shows risk (probability of constraint violation) and value for the starting state in regard to the policy computed with the algorithm of Neuneier and Mihatsch. Obviously, the algorithm is able to find the maximum-value policy yielding a zero deviation of $F_{spec}$, corresponding to choosing $F = F_{spec} = 0.8$ in all states, but the learning result is not sensitive to the risk parameter $\kappa$. The reason for this is that the worst-case and the best-case returns for the policy that always chooses the outflow 0.8 also correspond to

(a)                                                    (b)

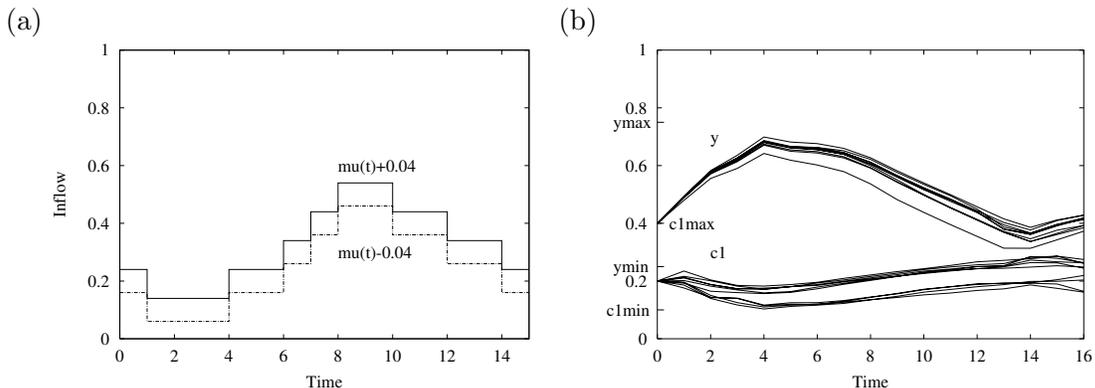

Figure 9: `RL-YC-CLC`: (a) $\mu(t) + 0.04$ and $\mu(t) - 0.04$ (profiles of the two mode means). (b) The tank level $y(t)$ and the concentration $c_1(t)$ for 10 example runs using the minimum risk policy.

0, which is the best return possible (implying a zero variance of the return). The approach of Neuneier and Mihatsch and variance-based approaches are therefore unsuited for the problem at hand.

## 8.2 The Problem `RL-YC-CLC` (Constraints for Tank Level and Concentrations)

In the following we will consider the full problem `RL-YC-CLC`. The two inflows $F_1$ and $F_2$ are assumed to have equal Gaussian distributions such that the distribution of the cumulative inflow $F_\Sigma(t) = F_1(t) + F_2(t)$ is described by the covariance matrix in (35) and the mean vector $\mu$ in (34); see also Figure 4(a).

In order to demonstrate the applicability of our approach to non-Gaussian distributions, we have chosen bimodal distributions for the inflow concentrations $c_1$ and $c_2$. The underlying assumption is that the upstream plants either all have an increased output, or all have a lower output, e.g. due to different hours or weekdays.

The distribution of the inflow concentration $c_{i,1}(t)$ is characterized by two Gaussian distributions with means

$$\mu(t) + (-1)^k 0.04,$$

where $k = 1, 2$ and $\sigma^2 = 0.0025$. The value of $k \in \{0, 1\}$ is chosen at the beginning of each run with equal probability for each outcome. This means that the overall mean value of $c_{i,1}(t)$ is given by $\mu(t)$. The profiles of the mean values of the modes can be found in Figure 9(a). $c_{i,2}$ is given as $c_{i,2}(t) = 1.0 - c_{i,1}(t)$. The minimum and maximum values for the concentrations $c_i(t)$ can be found in Table 1, and also in Figure 9(b). Note that the concentrations have to be controlled indirectly by choosing an appropriate outflow $F$.

The developing of the risk and the value of the starting state is shown in Figure 10. The resulting curves behave similar to that for the problem `RL-Y-CLC` depicted in Figure 5: both value and risk increase with $\xi$. It can be seen that the algorithm covers a relatively broad range of policies with different value-risk combinations.

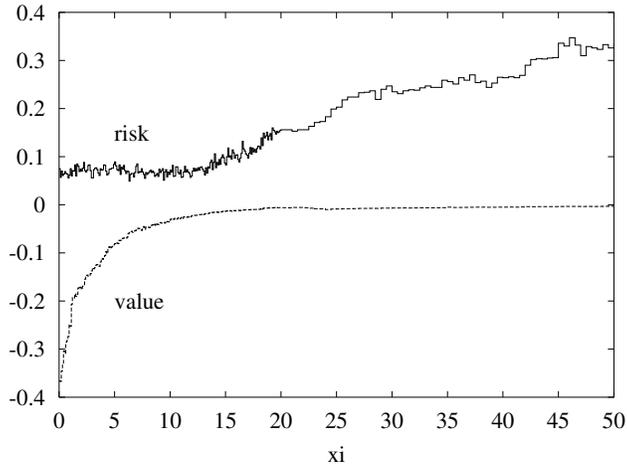

Figure 10: `RL-YC-CLC`: Estimated risk $\rho^{\pi^*_\xi}(x_0)$, value $V^{\pi^*_\xi}(x_0)$, and $V^{\pi^*_\xi}_\xi(x_0) = \xi V^{\pi^*_\xi}(x_0) - \rho^{\pi^*_\xi}(x_0)$ for different values of $\xi$.

For the minimum risk policy, the curves of the tank level $y$ and the concentration $c_1$ can be found in Figure 9(b). The bimodal characteristics of the substance 1 inflow concentrations are reflected by $c_1(t)$ (it holds $c_2(t) = 1 - c_1(t)$). The attainable minimum risk is 0.062. Increasing the weight $\xi$ leads to curves similar to that shown in the Figures 5 and 7. We assume that the minimum achievable risk can be decreased by the inclusion of additional variables, e.g. inflow rates and concentrations, and/or by the inclusion of past values as discussed in section 8.1.1. The treatment of a version with an action history is analogous to section 8.1.2. We therefore conclude presentation of the experiments at this point.

## 9. Conclusion

In this paper, we presented an approach for learning optimal policies with constrained risk for MDPs with error states. In contrast to other RL and DP approaches that consider risk as a matter of the variance of the return or of its worst outcomes, we defined risk as the probability of entering an error state.

We presented a heuristic algorithm that aims at learning good stationary policies that is based on a weighted formulation of the problem. The weight of the original value function is increased in order to maximize the return while the risk is required to stay below the given threshold. For a fixed weight and a finite state space, the algorithm converges to an optimal policy in the case of an undiscounted value function. For the case that the state space is finite, contains no cycles, and that $\gamma < 1$ holds, we conjecture the convergence of the learning algorithm to a policy, but assume that it can be suboptimal for the weighted formulation. If an optimal stationary policy exists for the weighted formulation, it is a feasible, but generally suboptimal solution for the constrained problem.

The weighted approach that is combined with an adaptation of $\xi$ is a heuristic for searching the space of feasible stationary policies of the original constrained problem, which to us seems relatively intuitive. We conjecture that better policies could be found by allowing state-dependent weights $\xi(x)$ with a modified adaptation strategy, and by extending the considered policy class.

We have successfully applied the algorithm to the control of the outflow of a feed tank that lies upstream of a distillation column. We started with a formulation as a stochastic optimal control problem with chance constraints, and mapped it to a risk-sensitive learning problem with error states (that correspond to constraint violation). The latter problem was solved using the weighted RL algorithm.

The crucial point in reformulation as an RL problem was the design of the state space. We found that the algorithm consistently performed better when state information was provided to the learner. Using the time and the action history resulted in very large state spaces, and a poorer learning performance. RBF networks together with sufficient state information facilitated excellent results.

It must be mentioned that the use of RL together with MLP or RBF network based function approximation suffers from the usual flaws: non-optimality of the learned network, potential divergence of the learning process, and long learning times. In contrast to an exact method, no a priori performance guarantee can be given, but of course an a posteriori estimate of the performance of the learned policy can be made. The main advantage of the RL method lies in its broad applicability. For the tank control task, we achieved very good results compared to those obtained through a (mostly) analytical approach.

For the cases $|X| > 1$ or $\gamma < 1$ further theoretical investigations of the convergence and more experiments are required. Preliminary experiments have shown that oscillations may occur in our algorithm, but the behavior tends to oscillate between sensible policies without getting too bad in-between although the convergence and usefulness of the policies remains an open issue.

Oscillations can be prevented by using a discounted risk that leads to an underestimation of the actual risk. The existence of an optimal policy and convergence of the learning algorithm for a fixed $\xi$ can be guaranteed in the case of a finite MDP. A probabilistic interpretation of the discounted risk can be given by considering $1 - \gamma$ as the probability of exiting from the control of the MDP (Bertsekas, 1995). The investigation of the discounted risk may be worthwhile in its own right. For example, if the task has long episodes, or if it is continuing, i.e. non-episodic, it can be more natural to give a larger weight to error states occurring closer to the current state.

We have designed our learning algorithm as an online algorithm. This means that learning is accomplished using empirical data obtained through interaction with a simulated or real process. The use of neural networks allows to apply the algorithm to processes with continuous state spaces. In contrast, the algorithm described by Dolgov and Durfee (2004) can only be applied in the case of a known finite MDP. Such a model can be obtained in the case of a continuous-state process by finding an appropriate discretization and estimating the state transition probabilities together with the reward function. Although such discretization prevents the application of Dolgov and Durfee's algorithm to `RL-Y-OLC`, where a 15-dimensional state space is encountered, it can probably be applied in the case of `RL-Y-OLC`. We plan to investigate this point in future experiments.

The question arises as to whether our approach can also be applied to stochastic optimal control problems with other types of chance constraints. Consider a conjunction of chance constraints

$$\mathbf{P}(C_0) \geq p_1, \ldots, \mathbf{P}(C_{N-1}) \geq p_{N-1}, \tag{37}$$

where each $C_t$ is a constraint system containing only variables at time $t$, and $p_t$ is the respective probability threshold. (37) requires an alternative RL formulation where the risk of a state only depends on the next reward, and where each time-step has its own $\omega_t$. The solution with a modified version of the RL algorithm is not difficult.

If each of the $C_t$ in (37) is allowed to be a constraint system over state variables depending on $t' \geq t$, things get more involved because several risk functions are needed for each state. We plan investigating these cases in the future.

**Acknowledgments** We thank Dr. Pu Li for providing the application example and for his helpful comments. We thank Önder Gencaslan for conducting first experiments during his master's thesis.